\documentclass{article}
\PassOptionsToPackage{table}{xcolor}
\usepackage[T1]{fontenc}

\usepackage{microtype}
\usepackage{graphicx}
\usepackage{subcaption}
\usepackage{booktabs} 

\usepackage{multirow}

\usepackage{hyperref}

\usepackage[accepted]{icml2026}

\usepackage{amsmath}
\usepackage{amssymb}
\usepackage{mathtools}
\usepackage{amsthm}
\usepackage{algorithm}
\usepackage{algpseudocode}

\usepackage[capitalize,noabbrev]{cleveref}
\usepackage{xcolor}
\usepackage{tikz}
\usetikzlibrary{positioning, arrows.meta, calc}

\theoremstyle{plain}

\theoremstyle{definition}

\theoremstyle{remark}

\usepackage[textsize=tiny]{todonotes}

\icmltitlerunning{Metal-Sci: A Scientific Compute Benchmark for Evolutionary LLM Kernel Search on Apple Silicon}

\begin{document}

\twocolumn[
  \icmltitle{Metal-Sci: A Scientific Compute Benchmark\\for Evolutionary LLM Kernel Search on Apple Silicon}



  \icmlsetsymbol{equal}{*}

  \begin{icmlauthorlist}
    \icmlauthor{Víctor Gallego}{yyy}
  \end{icmlauthorlist}

  \icmlaffiliation{yyy}{Komorebi AI Technologies, Madrid, Spain}

  \icmlcorrespondingauthor{Víctor Gallego}{victor.gallego@komorebi.ai}

  \icmlkeywords{LLM code generation, kernel optimization, evolutionary code search, scientific computing benchmark, Apple Silicon, Metal GPU programming}

  \vskip 0.3in
]



\printAffiliationsAndNotice{}  

\begin{abstract}
 We present \textsc{Metal-Sci}, a 10-task benchmark of scientific Apple  Silicon Metal compute kernels spanning six optimization regimes (stencils, all-pairs in $n$-body problems, multi-field Boltzmann, neighbor-list molecular dynamics, multi-kernel PDE, FFT). Each task ships a CPU reference, a  roofline-anchored fitness function, and a held-out generalization size. We pair the benchmark with a \emph{lightweight harness for automatic kernel search} that runtime-compiles each candidate, scores it against the roofline across multiple sizes, and feeds structured compile and per-size correctness diagnostics back to a frozen LLM driving a $(1{+}1)$ evolutionary loop. We report matched single-model sweeps of  Claude Opus~4.7, Gemini~3.1 Pro, and GPT-5.5 on M1~Pro: in-distribution self-speedups span $1.00\times$ to $10.7\times$. Beyond raw speedup, our central methodological claim is structural: the held-out gate scoring function $\Phi_\mathcal{T}$ (evaluated once at end-of-run on a configuration the agent never sees during search) functions as a cheap mechanical oversight primitive on this automatic search loop, catching e.g.\ an Opus \texttt{template <uint D>} HMC win that returns wrong samples at unseen dimensions, and a GPT FFT3D best that wins in-distribution at $2.95\times$ speedup but collapses to $0.23\times$ on a $256^3$ held-out cube, a silent regression that the in-distribution score alone cannot see. Code at\\
 \href{https://github.com/vicgalle/metal-sci-kernels}{github.com/vicgalle/metal-sci-kernels}.
\end{abstract}

\section{Introduction}

\begin{figure}[t]
\centering
\begin{tikzpicture}[
  himg/.style    = {inner sep=0pt, outer sep=0pt},
  nbox/.style    = {draw, rounded corners=2pt, line width=0.5pt,
                    minimum height=6.5mm, inner xsep=3pt, inner ysep=2pt,
                    align=center, font=\scriptsize\sffamily},
  llm/.style     = {nbox, fill=pink!18,   draw=pink!60!black,   minimum width=29mm},
  comp/.style    = {nbox, fill=black!7,   draw=black!50,        minimum width=29mm},
  disp/.style    = {nbox, fill=black!7,   draw=black!50,        minimum width=29mm},
  score/.style   = {nbox, fill=black!7,   draw=black!50,        minimum width=29mm},
  promote/.style = {nbox, fill=black!7,   draw=black!50,        minimum width=29mm},
  held/.style    = {nbox, fill=red!10,    draw=red!55!black,    minimum width=34mm,
                    minimum height=8mm,   font=\scriptsize\sffamily},
  fb/.style      = {nbox, fill=black!3,   draw=black!40,        font=\scriptsize\sffamily,
                    minimum height=22mm,  minimum width=29mm, align=left,
                    inner xsep=3.5pt,     inner ysep=3pt},
  arr/.style     = {-{Latex[length=1.6mm,width=1.6mm]}, line width=0.5pt},
  feed/.style    = {-{Latex[length=1.6mm,width=1.6mm]}, line width=0.5pt,
                    dashed, draw=gray!75!black},
  over/.style    = {-{Latex[length=1.6mm,width=1.6mm]}, line width=0.5pt,
                    dotted, draw=red!55!black},
  elbl/.style    = {font=\scriptsize\sffamily, inner sep=1pt},
  banner/.style  = {font=\scriptsize\sffamily\itshape, text=gray!55!black},
]
  \node[himg] (img) at (0,0) {\includegraphics[width=1.05\columnwidth]{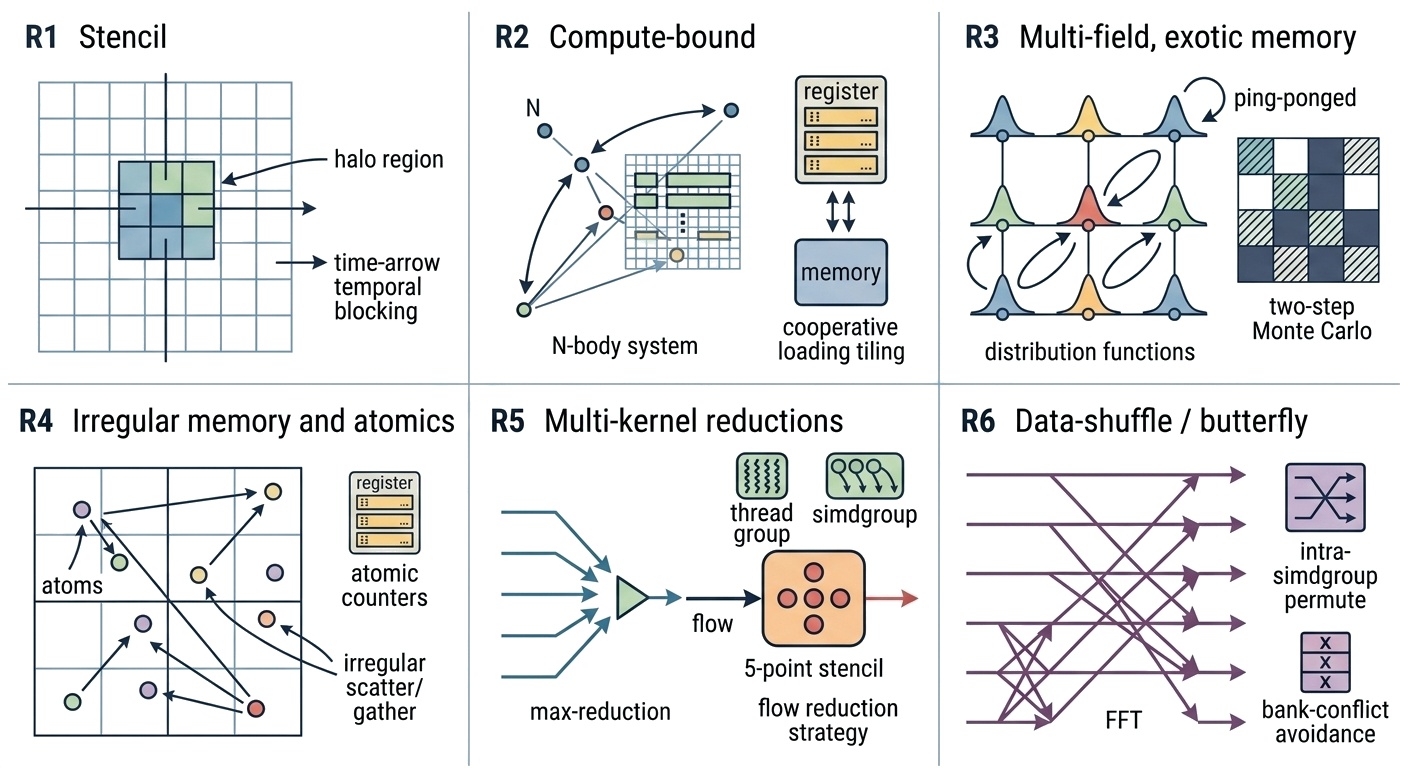}};

  \begin{scope}[shift={($(img.south)+(0,-0.30)$)}]
    \node[banner] at (0,-0.05) {kernel search harness loop --- iterated $K$ times per task $\mathcal{T}$};

    \node[llm]     (llm)   at (-1.70,-0.75) {Frozen LLM\\$\mathcal{M}\,(\mbox{prompt}_{\mathcal{T}},\,\mathcal{F}_{k-1})$};
    \node[comp]    (comp)  at (-1.70,-1.85) {Runtime compile};
    \node[disp]    (disp)  at (-1.70,-2.73) {Dispatch on $\Sigma_{\mathcal{T}}$ (3 sizes)};
    \node[score]   (score) at (-1.70,-3.70) {Score $S_{\mathcal{T}}(\kappa_k)$\\\scriptsize gmean $a_{\mathcal{T}}/c_{\mathcal{T}}$, gate $\chi_{\mathcal{T}}$};
    \node[promote] (promo) at (-1.70,-4.65) {$(1{+}1)$ promote\\$\kappa^{\star}_{k}$ if $S{>}S^{\star}$};
    \node[held]    (held)  at (-1.70,-5.95) {Held-out at $\sigma^{\star}_{\mathcal{T}}\!\to\!\Phi_{\mathcal{T}}(\kappa^{\star}_K)$\\(oversight, never in $\mathcal{F}_k$)};

    \node[fb] (fb) at (1.80,-2.85) {%
      \textbf{Feedback $\mathcal{F}_k$}\\
      \textbullet\ compile diag.\\
      \textbullet\ per-size $\chi_{\mathcal{T}}$\\
      \textbullet\ per-size $a_{\mathcal{T}}/c_{\mathcal{T}}$\\
      \textbullet\ short history};

    \draw[arr] (llm)   -- node[right,elbl]{kernel $\kappa_k$}     (comp);
    \draw[arr] (comp)  --                                   (disp);
    \draw[arr] (disp)  -- node[right,elbl]{} (score);
    \draw[arr] (score) --                                   (promo);

    \draw[over] (promo) -- node[right,elbl]{end-of-run, $\kappa^{\star}_K$} (held);

    \draw[feed] (promo.east) -| (fb.south);
    \draw[feed] (fb.north) |- (llm.east);
  \end{scope}
\end{tikzpicture}
\caption{The \textsc{Metal-Sci} benchmark.
\textbf{Top:} six optimization regimes (R1--R6), each stressing a structurally distinct GPU/memory bottleneck whose canonical recipe does not transfer to its neighbors (Sec.~\ref{sec:tasks}).
\textbf{Bottom:} the harness loop. A frozen LLM $\mathcal{M}$ emits a Metal source $\kappa_k$; the harness runtime-compiles it, dispatches across the in-distribution size configurations $\Sigma_{\mathcal{T}}$, and scores it against per-size roofline ceilings. The candidate becomes the new incumbent $\kappa^{\star}_{k}$ only when $S_{\mathcal{T}}$ strictly improves. Compile diagnostics and per-size $(\chi_{\mathcal{T}},\,a_{\mathcal{T}})$ flow back through a structured feedback packet $\mathcal{F}_k$ that primes the next iteration. The held-out evaluation $\Phi_{\mathcal{T}}$ at $\sigma^{\star}_{\mathcal{T}}$ runs once at end-of-run (Sec.~\ref{sec:harness}).}
\label{fig:overview}
\end{figure}

LLM code-search systems such as FunSearch \cite{funsearch}, AlphaEvolve \cite{novikov2025alphaevolve}, and recent kernel-generation work \cite{ouyang2025kernelbench} evaluate language models as
\emph{optimizers} of executable artifacts. The choice of artifact matters:
KernelBench targets PyTorch ML kernels (GEMM, attention, convolution) on
CUDA, where canonical implementations are heavily represented in
pretraining data and the optimization surface is dominated by tiling and
warp-level reductions.

Scientific computing exposes a different surface. For example, stencils are
bandwidth-bound and reward halo and temporal blocking. All-pairs
interactions are compute-bound and reward register blocking with
shared-memory cooperative loads. Lattice Boltzmann ping-pongs nine
distribution functions per cell with non-trivial pull-stream indexing.
Cell-list molecular dynamics touches irregular memory through atomic
counters. Hamiltonian Monte Carlo runs many chains in parallel where
register pressure scales with the problem dimension $d$. Grad-Shafranov
solvers chain a max-reduction with a variable-coefficient stencil. And 3D
FFTs are dominated by data-shuffle/butterfly patterns and twiddle-factor
caching rather than tiling. Each pattern stresses a distinct dimension of
the compiler/memory hierarchy, and canonical optimizations \cite{datta,nyland2007fast,schonherr2011multi,anderson2008general,govindaraju2008high} differ regime-to-regime.

We target the Apple Silicon's Metal compute pipeline. It is underrepresented
in CUDA-centric training data, as frontier models reliably mishandle
Metal-specific syntax such as the
\texttt{[[max\_total\_threads\_per\_threadgroup]]} kernel attribute, a simple
out-of-distribution generalization test. Its unified memory model removes
host-device copy plumbing, enabling sub-second compile-run-verify cycles
per candidate that are essential for fast evolutionary loops. And it
supports rich simdgroup intrinsics (\texttt{simd\_max},
\texttt{simd\_broadcast}) and threadgroup-memory tiling, giving the LLM a
non-trivial optimization surface.

\paragraph{Contributions.} (i) A 10-task scientific compute benchmark over
six optimization regimes, each with a CPU reference, roofline-anchored
fitness function, and multi-size generalization gate; (ii) a runtime-compiled harness that exposes compile errors, correctness diagnostics, and per-size GPU timing back to the LLM; and (iii) three matched single-model sweeps (Claude Opus~4.7, Gemini~3.1~Pro, and GPT-5.5) on M1~Pro that operationalize the held-out gate $\Phi_\mathcal{T}$ as a cheap mechanical oversight primitive on the coding agent: $\Phi_\mathcal{T}$ is never folded into any feedback packet seen by the LLM during search and catches confidently-wrong outputs (silent correctness violations and silent regressions) that the in-distribution score $S_\mathcal{T}$ alone licenses.

\paragraph{Related work.} Existing LLM kernel-generation benchmarks, such as KernelBench~\cite{ouyang2025kernelbench},
TritonBench~\cite{li2025tritonbench},
BackendBench~\cite{saroufim2025backendbench},
MultiKernelBench~\cite{wen2025multikernelbench},
NPUEval~\cite{kalade2025npueval} and
KernelCraft~\cite{nie2026kernelcraft},
target ML operators on CUDA or accelerator backends and score speedup
against a vendor or compiler baseline (Table~\ref{tab:related}).
\textsc{Metal-Sci} differs on three axes that drive the harness
design: \emph{(i)}~scientific compute spanning six structurally
distinct optimization regimes (R1--R6, Sec.~\ref{sec:tasks}) whose
canonical recipes \cite{datta,nyland2007fast,schonherr2011multi,anderson2008general,govindaraju2008high}
do not transfer between regimes; \emph{(ii)}~a per-chip roofline
score (decoupled from any reference implementation) with a held-out
size configuration gate $\Phi_\mathcal{T}$ the agent never sees during search;
and \emph{(iii)}~Apple Silicon Metal, which is structurally underrepresented
in CUDA-centric pretraining and a real OOD test on Metal-grammar
idioms (\texttt{[[max\_total\_threads\_per\_threadgroup]]} attribute
placement, \texttt{half} as a reserved fp16 keyword, no C++ lambdas).
The broader paradigm of LLMs as optimizers in evolutionary code
search is established by FunSearch~\cite{funsearch} and
AlphaEvolve~\cite{novikov2025alphaevolve} and specialised to CUDA ML
kernels by AI~CUDA~Engineer~\cite{lange2025towards} and
EvoEngineer~\cite{guo2025evoengineer}; App.~\ref{app:related} expands
this related work.

\begin{table*}[!t]
\caption{\textsc{Metal-Sci} versus existing LLM kernel-generation
benchmarks across the axes that drive the design of our harness.
\emph{Domain}: ML = neural-network operators (GEMM, attention,
convolution, normalization, activation); HPC = scientific compute
(stencil, $N$-body, LBM, MD, MCMC, PDE solver, FFT). \emph{Score basis}:
what \emph{achieved} is normalized against. \emph{Multi-size}: whether
the score requires generalization across problem sizes. \emph{Search}:
the loop structure exposed to the LLM. Within the ML row, all listed
benchmarks span a single optimization regime (matmul-style, with
norm/activation epilogues); \textsc{Metal-Sci} spans six structurally
distinct regimes (R1--R6 of Section~\ref{sec:tasks}).}
\label{tab:related}
\centering
\footnotesize
\setlength{\tabcolsep}{4pt}
\resizebox{\textwidth}{!}{%
\begin{tabular}{l l l l l l}
\toprule
Benchmark & Target & Domain & Score basis & Multi-size & Search \\
\midrule
KernelBench \cite{ouyang2025kernelbench}       & CUDA                      & ML  & PyTorch speedup      & ---            & single-shot \\
TritonBench \cite{li2025tritonbench}      & Triton DSL                & ML  & speedup\,+\,code-sim & ---            & single-shot \\
BackendBench \cite{saroufim2025backendbench}     & PyTorch backend           & ML  & correctness only     & ---            & single-shot \\
MultiKernelBench \cite{wen2025multikernelbench} & CUDA / Ascend\,C / Pallas & ML  & compiler speedup     & ---            & multi-turn \\
NPUEval \cite{kalade2025npueval}          & AIE C++ (AMD NPU)         & ML  & compiler speedup     & ---            & tool-use \\
KernelCraft \cite{nie2026kernelcraft}      & accelerator asm           & ML  & compiler speedup     & cfg.\ vars     & agentic tool-use \\
\midrule
\textbf{\textsc{Metal-Sci} (ours)} & \textbf{Apple Metal MSL} & \textbf{HPC} & \textbf{\% of roofline} & \textbf{3 in\,+\,1 held-out} & \textbf{evolutionary $(1{+}1)$} \\
\bottomrule
\end{tabular}%
}
\end{table*}

\paragraph{Background and vocabulary.}
\textit{Apple GPU terms.} A \emph{kernel} is a
\texttt{.metal}-source GPU function launched from the CPU; a
\emph{threadgroup} (TG) is Apple's name for a CUDA thread block:
cooperating threads sharing a scratchpad ``threadgroup memory''; a
\emph{simdgroup} is the 32-thread SIMD lane group inside a threadgroup
(Apple's name for a CUDA warp); the \emph{System-Level Cache} (SLC) is
the CPU/GPU-shared last-level cache (${\sim}24$\,MB on M1~Pro) sitting
between the on-chip caches and DRAM.
\textit{Roofline.} A kernel's roofline~\cite{williams2009roofline}
ceiling is the per-chip throughput upper bound implied by its arithmetic
intensity (FLOPs per byte transferred). Kernels with low intensity are
\emph{bandwidth-bound} and cap at peak DRAM bandwidth (GB/s);
high-intensity ones are \emph{compute-bound} and cap at peak FP32
throughput (GFLOPS). We score candidates as a fraction of this hardware
ceiling rather than against any hand-written baseline.
\textit{Evolution strategy} $(\mu{=}1{+}\lambda{=}1)$, also called
$(1{+}1)$ \cite{beyer2002evolution}, denotes the simplest evolution strategy: one parent, one
mutated child per iteration; the child replaces the parent iff it
scores higher.

\section{Benchmark tasks}
\label{sec:tasks}

\begin{table*}[t]
\caption{The 10 \textsc{Metal-Sci} tasks. ``Optimization lever''
names the dominant move an LLM must reach for in that regime.
$N_x{\times}N_y$ grids written as $N^2$ when square; cube edges as
$N^3$. \texttt{saxpy} is a bandwidth smoke-test outside the regime
structure, used to validate the harness.}
\label{tab:tasks}
\centering
\footnotesize
\setlength{\tabcolsep}{4pt}
\resizebox{\textwidth}{!}{%
\begin{tabular}{l l l l l}
\toprule
Regime & Task & Optimization lever & In-distribution size configurations & Held-out \\
\midrule
\rowcolor{blue!4}    R1 stencil      & \texttt{heat2d}   & halo, temporal blocking                       & $\{256,512,1024\}^2$ & $768^2$ \\
\rowcolor{blue!4}                    & \texttt{wave3d}   & 2.5D blocking, register pressure              & $\{64,160,192\}^3$   & $128^3$ \\
\rowcolor{orange!5} R2 compute      & \texttt{nbody}    & register tiling, threadgroup cooperative load & $N{\in}\{256,1024,2048\}$ & $512$ \\
\rowcolor{orange!5}                 & \texttt{hmc}      & per-thread state vs.\ register file           & $(d{,}K){\in}\{(8{,}16K){,}(16{,}4K){,}(32{,}1K)\}$ & $(24{,}2K)$ \\
\rowcolor{green!5}  R3 multi-field  & \texttt{lbm}      & SoA layout, BGK algebraic fold                & $\{64,128,256\}^2$   & $192^2$ \\
\rowcolor{green!5}                  & \texttt{ising}    & checkerboard MC, byte-exact verify            & $\{256,1024,2048\}^2$ & $1536^2$ \\
\rowcolor{violet!5}R4 atomics                    & \texttt{lj}       & cell-list scatter, atomic contention          & $N{\in}\{1.7,4.1,10.6\}{\rm K}$ & $2744$ \\
\rowcolor{yellow!9}R5 multi-kernel               & \texttt{gradshaf} & in-kernel reduction $+$ var-coef stencil      & $\{65,257,513\}^2$   & $129^2$ \\
\rowcolor{cyan!6}R6 butterfly                    & \texttt{fft3d}    & TG bank conflicts, mixed-radix, \texttt{simd\_shuffle} & $\{32,64,128\}^3$ & $256^3$ \\
\midrule
\rowcolor{gray!5}(smoke)                         & \texttt{saxpy}    & DRAM saturation                               & $\{1,16,64\}{\rm M}$ & $4{\rm M}$ \\
\bottomrule
\end{tabular}%
}
\end{table*}

\textsc{Metal-Sci} packages 10 tasks into six \emph{optimization
regimes} (R1--R6, Fig.~\ref{fig:overview}\,top, Table~\ref{tab:tasks}). The choice of regimes is
the benchmark's central design decision: each regime stresses a
structurally distinct dimension of the GPU/memory hierarchy, and its
canonical recipe~\cite{datta,nyland2007fast,schonherr2011multi,anderson2008general,govindaraju2008high}
does not transfer to its neighbours. A halo-blocking move that wins
R1 is useless on R2 (register tiling), R4 (atomic contention), or R6
(intra-simdgroup butterflies). For an LLM, this is the structural
reason recall alone cannot solve the suite: there is no template that
wins everywhere, so the model has to actually \emph{recognise the
regime} from the kernel seed and reach for the right lever.

Each task ships (a)~a Metal seed kernel, (b)~a CPU reference with
task-specific tolerance, to check correctness, (c)~three in-distribution size configurations
plus one held-out size configuration, and (d)~a per-size roofline ceiling in
GFLOPS (compute-bound) or GB/s (bandwidth-bound). Per-task equations,
ceilings, and verification details are deferred to App.~\ref{app:tasks};
below we name the lever each regime tests.

\paragraph{R1 -- Regular stencils.} Bandwidth-bound updates over a
structured grid: a 5-point heat-equation stencil (8\,B/cell) and a
7-point leapfrog wave equation (12\,B/cell). The lever is halo
handling, marching-axis choice, and 2.5D temporal blocking.
\texttt{wave3d} doubles as a NaN trap, that is, a sign or indexing error
compounds over many leapfrog steps (e.g. 10 of Opus's 13 correctness
fails in Sec.~\ref{sec:results} land here).

\paragraph{R2 -- Compute-bound.} $O(N^2)$ pair sums from $n$-body simulations (\texttt{nbody})
and $O(d^2)$ matvec inside an $L$-step leapfrog integration (\texttt{hmc}), both
running ${\sim}20$ FLOPs per memory transaction so the ceiling is
peak FP32 GFLOPS. The lever is register tiling and threadgroup
cooperative loads. \texttt{hmc} additionally probes the
register-pressure boundary and verifies correctness
statistically (sample mean and Frobenius covariance error vs.\ the
target Gaussian).

\paragraph{R3 -- Multi-field, exotic memory.} \texttt{lbm} is a D2Q9
Lattice Boltzmann (pull-stream $+$ BGK collision) with nine
distribution fields per cell at 72\,B/cell traffic; the lever is
SoA layout, push/pull streaming choice, and algebraic factorisations
of the BGK relaxation (App.~\ref{app:code} dissects an FMA fold
Opus discovered). \texttt{ising} is 2D Ising checkerboard Metropolis
Monte Carlo at 2\,B/site; a precomputed accept-probability table and a
counter-based Murmur-fmix32 PRNG yield bit-exact CPU/GPU agreement,
so verification reduces to byte-equality on the spin array.

\paragraph{R4 -- Irregular memory and atomics.} Lennard-Jones molecular dynamics
with a cell-list spatial hash. Three kernels per step
(\texttt{clear\_cells} / \texttt{build\_cells} / \texttt{step});
\texttt{build\_cells} is an atomic scatter onto per-cell occupancy
counters and the force kernel walks 27 neighbor cells with
minimum-image periodic wrap. The lever is load balancing under
uneven cell occupancy and atomic-contention mitigation.

\paragraph{R5 -- Multi-kernel reductions.} Picard iteration for the
Grad-Shafranov fixed-boundary plasma equilibrium. Each outer step
dispatches a max-reduction
followed by a variable-coefficient 5-point stencil with a nonlinear
source. The lever is the choice of in-kernel reduction strategy 
(single-threadgroup vs.\ simdgroup-tree) and how cleanly the two
kernels compose across the dispatch boundary.

\paragraph{R6 -- Data-shuffle / butterfly.} 3D complex-to-complex
forward Fast Fourier Transform (FFT), dispatched as three per-axis 1D FFTs with two
ping-ponged buffers. Unlike R1/R2, the optimization surface is
\emph{data movement} rather than arithmetic: bit-reversal vs.\
Stockham auto-sort, twiddle-factor caching, mixed-radix (radix-4,
radix-8) butterflies for fewer barriers, and intra-simdgroup
permutes via \texttt{simd\_shuffle\_xor}. The per-axis stride
asymmetry (stride 1 along $x$ vs.\ $N$/$N^2$ along $y,z$) makes
threadgroup-memory bank-conflict avoidance a separate sub-lever.

\section{Harness Design}
\label{sec:harness}

The harness we propose closes an evolutionary loop (Fig.~\ref{fig:overview}\,bottom) around a frozen LLM: each
  iteration runtime-compiles a candidate Metal source, dispatches it
  across the task's in-distribution size configurations, scores the result against the
  per-chip roofline, and packs compile diagnostics, per-size
  correctness, and per-size throughput into a structured feedback
  packet $\mathcal{F}_k$ that primes the next iteration. We adopt two design choices: \emph{(i)}~runtime compilation inside
  a Python process, so each $(1{+}1)$ step costs seconds rather than
  minutes and the agent can iterate on the same kernel dozens of times
  per task; and \emph{(ii)}~a held-out score $\Phi_{\mathcal{T}}$ at an unseen size configuration, computed once at end-of-run and never folded into
  any $\mathcal{F}_k$ the LLM sees during search, to also test for generalization. 

\paragraph{Compile and dispatch.} The harness uses PyObjC's Metal bindings
and runtime-compiles \texttt{.metal} source via
\texttt{MTLDevice.newLibraryWithSource}, avoiding the offline
\texttt{xcrun metal} toolchain; compile errors are returned as structured
strings to the LLM. Buffers are allocated in unified memory. All dispatches for a
multi-size run share one \texttt{MTLCommandBuffer}, and timings come from
\texttt{GPUEndTime}\,$-$\,\texttt{GPUStartTime} (3 warmup, 10 timed,
median reported). The chip is detected from \texttt{sysctl} and looked up
in a per-family table (M1 through M4) for peak FP32 GFLOPS and DRAM
bandwidth.

\paragraph{Notation.} Each task ships a seed kernel
$\kappa_{\mathcal{T}}$ (Metal source), three in-distribution size
configurations $\Sigma_{\mathcal{T}}{=}\{\sigma_1,\sigma_2,\sigma_3\}$, a held-out
size $\sigma^{\star}_{\mathcal{T}}{\notin}\Sigma_{\mathcal{T}}$, and a
per-size roofline $c_{\mathcal{T}}(\sigma)$ in GFLOPS or GB/s. Evaluating a
candidate $\kappa$ at size config $\sigma$ produces a correctness flag
$\chi_{\mathcal{T}}(\kappa,\sigma){\in}\{0,1\}$ (CPU reference within
tolerance or not) and an achieved throughput
$a_{\mathcal{T}}(\kappa,\sigma)$; denote
$f_{\mathcal{T}}(\kappa,\sigma){=}a_{\mathcal{T}}(\kappa,\sigma)/c_{\mathcal{T}}(\sigma)$
for the per-size fraction-of-ceiling.

\paragraph{Scoring.} The in-distribution score is the geometric mean of
$f_{\mathcal{T}}$ over in-distribution size configurations, gated on correctness on \emph{every}
size:
\begin{equation*}
S_{\mathcal{T}}(\kappa) \;=\;
\Big(\textstyle\prod_{\sigma\in\Sigma_{\mathcal{T}}} f_{\mathcal{T}}(\kappa,\sigma)\Big)^{1/|\Sigma_{\mathcal{T}}|}
\,\cdot\,
\textstyle\prod_{\sigma\in\Sigma_{\mathcal{T}}}\chi_{\mathcal{T}}(\kappa,\sigma).
\end{equation*}
The hard gate ($S_{\mathcal{T}}{=}0$ on any tolerance failure) prevents
trading correctness for speed; the gmean across sizes discourages overfit
to one regime. The held-out gate, run only on the run's incumbent at
end-of-run, is the analogous quantity at the unseen size configuration:
$\Phi_{\mathcal{T}}(\kappa){=}f_{\mathcal{T}}(\kappa,\sigma^{\star}_{\mathcal{T}}){\cdot}\chi_{\mathcal{T}}(\kappa,\sigma^{\star}_{\mathcal{T}})$.
$S_{\mathcal{T}}$ is the agent's optimization target; $\Phi_{\mathcal{T}}$
is the external oversight signal it never sees.

\paragraph{Evolution loop.} A frozen LLM $\mathcal{M}$ acts as a kernel
synthesizer: given a task-spec system prompt $p_{\mathcal{T}}$ and a
feedback packet $\mathcal{F}_k$ summarizing the previous candidate, the
incumbent, and a short per-iteration history, $\mathcal{M}$ emits the
next Metal source $\kappa_{k+1}$ (with
$\kappa_0{=}\kappa_{\mathcal{T}}$). The harness compiles it, dispatches
across $\Sigma_{\mathcal{T}}$, and scores it; the strict $(1{+}1)$ rule
replaces the incumbent only when the new candidate scores strictly
higher under $S_{\mathcal{T}}$. We formalize this
compile--evaluate--promote $(1{+}1)$ loop in
Alg.~\ref{alg:loop} (App.~\ref{app:algorithm}). Compile, pipeline,
and per-size correctness errors are returned inside $\mathcal{F}_{k+1}$ as
structured strings (the violating size, the error metric and value, and
the compiler diagnostic if any). After $K$ iterations the run terminates;
both $S_{\mathcal{T}}(\kappa^{\star}_K)$ and the held-out
$\Phi_{\mathcal{T}}(\kappa^{\star}_K)$ are reported. 

\begin{table*}[t]
\caption{Evolutionary kernel refinement sweeps on Apple M1~Pro
(Opus = claude-opus-4-7, Gemini = gemini-3.1-pro-preview,
GPT = gpt-5.5).
\emph{In-dist.\ $\times$} = best / seed, gmean over three in-distribution size configurations. \emph{Held-out frac-of-ceiling} = achieved/ceiling at
the unseen size, measured in a single fresh session for all three
models; \emph{held-out $\times$} = best / seed at that size config.
\textbf{Bold} in the in-distribution and held-out speedup columns
marks meaningful improvements ($\geq 1.05\times$).}
\label{tab:results}
\centering
\footnotesize
\setlength{\tabcolsep}{3pt}
\resizebox{\textwidth}{!}{%
\begin{tabular}{l ccc ccc ccc l}
\toprule
       & \multicolumn{3}{c}{In-dist.\ $\times$} & \multicolumn{3}{c}{Held-out frac-of-ceiling} & \multicolumn{3}{c}{Held-out $\times$} & \\
\cmidrule(lr){2-4}\cmidrule(lr){5-7}\cmidrule(lr){8-10}
Task     & Opus & Gemini & GPT & Opus    & Gemini & GPT & Opus            & Gemini         & GPT        & Outcome \\
\midrule
saxpy    & \textbf{1.25} & 1.00          & 1.01          & 93\%    & 78\%   & 78\%    & \textbf{1.17}   & 0.98            & 0.98            & saturated \\
heat2d   & 1.00          & 1.03          & 1.00          & 84\%    & 98\%   & 80\%    & 0.86            & 1.01            & 0.82            & saturated \\
wave3d   & \textbf{1.26} & 1.00          & 1.00          & 97\%    & 87\%   & 96\%    & 1.00            & 0.90            & 0.99            & saturated \\
ising    & \textbf{1.13} & 1.00          & \textbf{1.09} & 11\%    & 12\%   & 10\%    & 0.94            & 0.99            & 0.88            & flat \\
fft3d    & 1.03          & \textbf{1.19} & \textbf{2.95} & 42\%    & 45\%   & 8.5\%   & \textbf{1.12}   & \textbf{1.20}   & {\textcolor{purple}{0.23}}            & \textbf{GPT silent regression} \\
nbody    & \textbf{2.83} & \textbf{2.00} & \textbf{2.19} & 1.6\%   & 2.0\%  & 1.8\%   & \textbf{1.24}   & \textbf{1.50}   & \textbf{1.37}   & generalizes \\
gradshaf & \textbf{1.89} & \textbf{2.89} & \textbf{1.93} & 5.4\%   & 7.7\%  & 5.0\%   & \textbf{2.05}   & \textbf{2.91}   & \textbf{1.86}   & generalizes \\
lj       & \textbf{1.77} & \textbf{1.98} & \textbf{1.62} & 0.31\%  & 0.47\% & 0.34\%  & \textbf{1.24}   & \textbf{1.87}   & \textbf{1.34}   & generalizes \\
lbm      & \textbf{1.46} & \textbf{1.06} & \textbf{1.33} & 79\%    & 93\%   & 82\%    & 0.97            & \textbf{1.16}   & 1.01            & tied at $192^2$ \\
hmc      & \textbf{10.6} & \textbf{10.7} & \textbf{7.19} & ---     & 9.7\%  & 10.2\%  & {\textcolor{purple}{FAIL}}            & \textbf{17.6}   & \textbf{18.6}   & \textbf{Opus wrong at $d{=}24$}; Gemini, GPT generalize \\
\bottomrule
\end{tabular}%
}
\end{table*}

\section{Experiments}
\label{sec:results}

\begin{figure*}[!t]
\centering
\includegraphics[width=\textwidth]{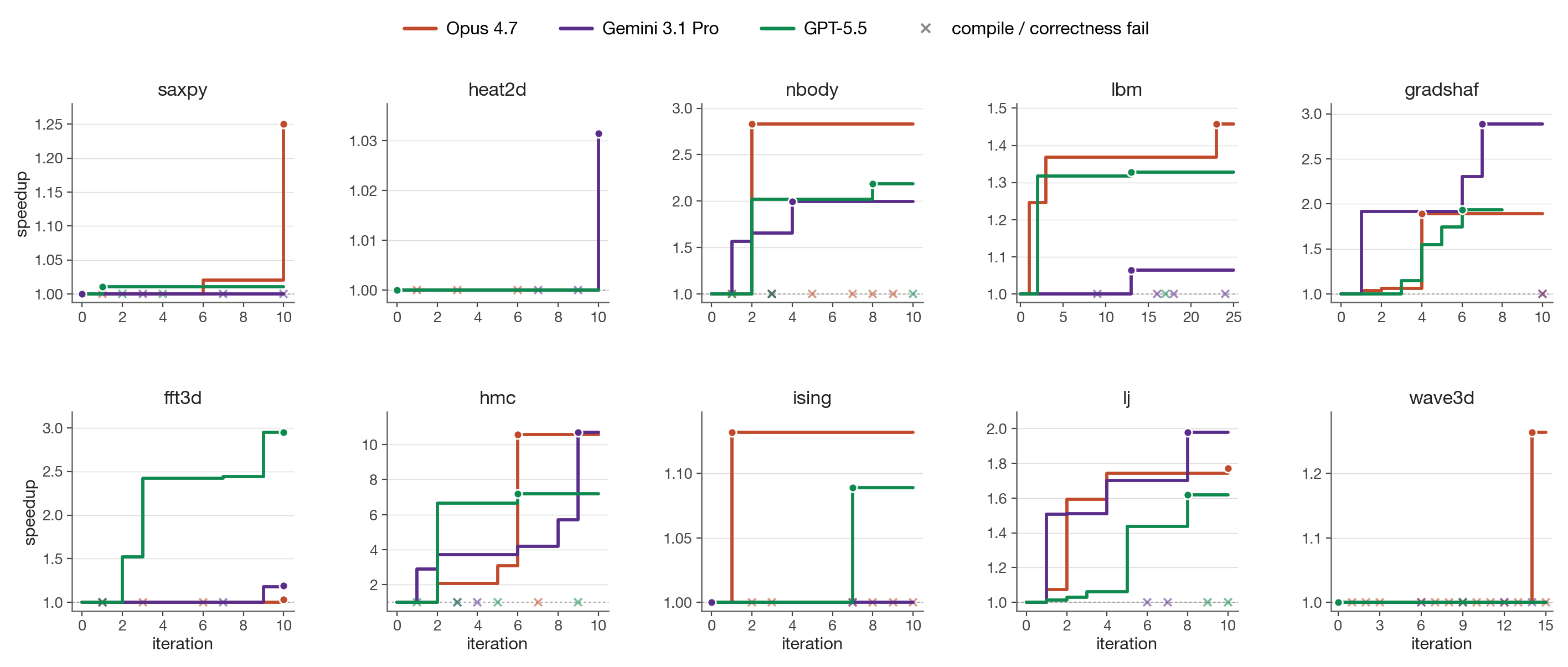}
\caption{Per-task running self-speedup (best-so-far / seed) versus
iteration, Opus 4.7 (orange) vs Gemini 3.1 Pro (purple) vs GPT-5.5
(green). Filled circles mark the iteration that achieved the
final best; each \texttt{x} along $y{=}1$ is a candidate that
compiled or ran wrong (the incumbent is unchanged). The visible
counts make the silent-correctness story concrete: Opus emits more
failures than Gemini or GPT at every task with non-trivial
search (\texttt{nbody}, \texttt{hmc}, \texttt{ising}, \texttt{lj}, \texttt{wave3d}). The plot also justifies
the asymmetric per-task budgets: on most tasks (\texttt{saxpy}, \texttt{heat2d},
\texttt{nbody}, \texttt{gradshaf}, \texttt{hmc}, \texttt{ising}, \texttt{lj}) the incumbent stops moving by
iter~$\sim$8 for all three models, but \texttt{lbm}'s Opus run plateaus at
$1.36\times$ from iter~3 through iter~22 and only breaks through
to $1.46\times$ at iter~23 (the BGK fold + pinned threadgroup of
App.~\ref{app:code}); \texttt{wave3d}'s Opus run shows the same shape with
the lift at iter~14 ($1.26\times$). A 10-iter budget would have
missed both Opus exemplars. GPT's \texttt{fft3d} climbs monotonically to $2.95\times$, and is exactly the win that fails to
generalize on the held-out gate.}
\label{fig:convergence}
\end{figure*}

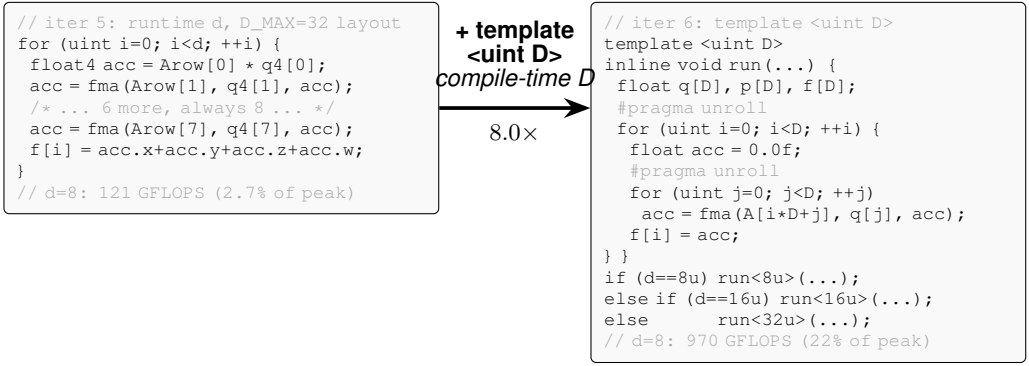
\begin{figure*}[!t]
\centering
\begin{tikzpicture}[
  code/.style={draw, rounded corners=2pt, fill=gray!5,
               inner sep=5pt, align=left, font=\scriptsize\ttfamily,
               text width=54mm, anchor=north west},
  arr/.style={-{Stealth[length=3.5mm]}, line width=1.4pt},
]
\node[code] (l) at (0,0) {%
\textcolor{gray!70}{// iter 5: runtime d, D\_MAX=32 layout}\\
for (uint i=0; i<d; ++i) \{\\
\ \ float4 acc = Arow[0] * q4[0];\\
\ \ acc = fma(Arow[1], q4[1], acc);\\
\ \ \textcolor{gray!70}{/* ... 6 more, always 8 ... */}\\
\ \ acc = fma(Arow[7], q4[7], acc);\\
\ \ f[i] = acc.x+acc.y+acc.z+acc.w;\\
\}\\
\textcolor{gray!70}{// d=8: 121 GFLOPS (2.7\% of peak)}};

\node[code] (r) at ($(l.north east)+(20mm,0)$) {%
\textcolor{gray!70}{// iter 6: template <uint D>}\\
template <uint D>\\
inline void run(...) \{\\
\ \ float q[D], p[D], f[D];\\
\ \ \textcolor{gray!70}{\#pragma unroll}\\
\ \ for (uint i=0; i<D; ++i) \{\\
\ \ \ \ float acc = 0.0f;\\
\ \ \ \ \textcolor{gray!70}{\#pragma unroll}\\
\ \ \ \ for (uint j=0; j<D; ++j)\\
\ \ \ \ \ \ acc = fma(A[i*D+j], q[j], acc);\\
\ \ \ \ f[i] = acc;\\
\} \}\\
if (d==8u) run<8u>(...);\\
else if (d==16u) run<16u>(...);\\
else \ \ \ \ \ \ \ \ \ \ run<32u>(...);\\
\textcolor{gray!70}{// d=8: 970 GFLOPS (22\% of peak)}};

\coordinate (lm) at ($(l.east|-l.center)$);
\coordinate (rm) at (lm-|r.west);
\node[font=\footnotesize\sffamily, align=center, above=1mm of $(lm)!0.5!(rm)$]
  (lab) {\textbf{+ template}\\[-1pt]\textbf{<uint D>}\\[-1pt]\textit{compile-time D}};
\node[font=\footnotesize\sffamily\bfseries, below=1mm of $(lm)!0.5!(rm)$] {$8.0\times$};
\draw[arr] (lm) -- (rm);
\end{tikzpicture}
\caption{Paradigmatic candidate evolution: \texttt{hmc}, Opus~4.7, iter~5\,$\rightarrow$\,6.
Both Opus and Gemini independently arrive at this structural change.
The lever is one declaration: \texttt{template <uint D>}
with runtime-dispatched instantiations on $d$. Iter~5 had a
manually-unrolled float4 inner loop against a fixed $D_{\max}{=}32$ layout
(over-computing at $d{=}8$, plus a 4-way horizontal sum); iter~6 sizes
\texttt{q,p,f} exactly to $D$ so both loops fully unroll into scalar FMAs ---
moving $d{=}8$ from 121 to 970\,GFLOPS in one iteration after five had
stalled at 2--4\% of peak. Opus enumerated $D{\in}\{8,16,32\}$ and broke
correctness at the held-out $d{=}24$; Gemini kept a runtime-$d$ fallback
and generalized cleanly.}
\label{fig:hmc-evol}
\end{figure*}

We run three matched single-model sweeps on Apple M1~Pro (4500\,GFLOPS, 200\,GB/s) ---
Claude Opus 4.7, Gemini 3.1 Pro, and
GPT-5.5 --- over the ten tasks at the same per-task
iteration budget (10 each except lbm at 25 and wave3d at 15\footnote{the asymmetry tracks where the incumbent kept moving past iter~10
in pilot runs and is justified by
Fig.~\ref{fig:convergence}}),
$\mu{=}1{+}\lambda{=}1$, no human prompt intervention.
Table~\ref{tab:results} reports each model's in-distribution self-speedup
(best / seed, gmean over three in-distribution size configuration) and a held-out evaluation
in which the unmodified seed and each incumbent best are run on a single
unseen problem size configuration declared by the task spec (Sec.~\ref{sec:tasks}).
The held-out columns are remeasured in a single fresh session for all
three models so the absolute fraction-of-ceiling numbers stay
apples-to-apples; we observed that single-size held-out fractions can
shift by tens of percentage points across sessions due to GPU thermal
state and SLC residency, so we anchor on the within-session ratios.

\paragraph{In-distribution results.} Self-speedups span $1.00\times$ to
$10.7\times$. The \texttt{hmc} step (Fig.~\ref{fig:hmc-evol}) is the high end for Opus-4.7 and Gemini-3.1-Pro: each independently introduces a
\texttt{template <uint D>} worker dispatched on runtime $d$ that takes
$d{=}8$ from $\sim$120 to $\sim$970\,GFLOPS by enabling full unroll
of the inner $A\,q$ matvec, and the two top scores agree to within
1.4\% (Opus 0.0932 vs Gemini 0.0870); GPT-5.5 reaches the same regime more
cautiously at $7.2\times$ (GPT 0.0634). Outside \texttt{hmc} the models split:
Opus wins on \texttt{saxpy} ($1.25$), \texttt{nbody} ($2.83$), \texttt{lbm} ($1.46$), and
\texttt{wave3d} ($1.26$); Gemini wins on \texttt{gradshaf} ($2.89$) and \texttt{lj} ($1.98$);
GPT wins \texttt{fft3d} outright at $2.95\times$ (vs $1.19$ Gemini,
$1.03$ Opus), the largest in-distribution gap any single model opens
on the suite, and is competitive on \texttt{nbody} ($2.19$), \texttt{gradshaf} ($1.93$),
\texttt{ising} ($1.09$), and \texttt{lbm} ($1.33$). The Opus--Gemini split correlates
with the type of optimization lever: ``tune the same algorithm
tighter'' (BW saturation, BGK fold, leapfrog ILP) favors Opus,
while ``find a different algorithm'' (simdgroup-tree reduction,
twiddle caching, neighbor-list reorganization) favors Gemini (see
App.~\ref{app:code} for code-level diffs on \texttt{lbm} and \texttt{fft3d}, plus
GPT's \texttt{fft3d} direct-DFT fallback and \texttt{hmc} defensive
enumeration). GPT
sits closer to Gemini in temperament: it explores aggressive
restructurings, which on \texttt{fft3d} pays off in-distribution but, as the
held-out column makes visible, at the cost of a sharp overfit
(see next paragraph).
Saturated tasks (\texttt{saxpy},
\texttt{heat2d}, \texttt{wave3d}) sit above 78\% of effective DRAM ceiling on the seed;
\texttt{saxpy} and \texttt{heat2d} primarily validate the harness, leaving the search
loop little to do, and \texttt{heat2d} shows the score is a \emph{hard}
signal: on both Opus and GPT zero candidates strictly dominated
the seed across all three sizes and the incumbent stayed at iter~0.
Across all three sweeps, Gemini had \emph{zero} correctness failures
across all its candidates; Opus had 13
(notably \texttt{wave3d} 10/15: multi-step leapfrog amplifies any sign or
indexing error into NaN); GPT had 2 across all candidates,
the second-lowest correctness-failure rate of the three.

\paragraph{Held-out generalization is sharper than in-distribution, and the three models overfit asymmetrically.}
We distinguish two main result populations. (i)~\emph{Generalizes}: \texttt{nbody}, \texttt{gradshaf},
\texttt{lj}, and (for Opus and Gemini) \texttt{fft3d}. \texttt{gradshaf} is the standout where all
three models extrapolate cleanly ($2.05\times$ Opus, $2.91\times$ Gemini,
$1.86\times$ GPT); on \texttt{lj} only Gemini exceeds its in-distribution
speedup ($1.87\times$ at $N{=}2744$); Opus and GPT hold partial
gains ($1.24\times$ and $1.34\times$ respectively).
(ii)~\emph{Overfits}, with the most consequential disagreements
between the three models. \texttt{hmc} is the sharpest correctness
case: Opus's template specialization dispatches
\texttt{if (d==8) run<8>() ... else run<32>()}, so $d{=}24$ lands
in the $D{=}32$ branch, per-thread \texttt{q[32], p[32]} and the
unrolled matvec process 32 entries against 24-entry data, and
sample covariance lands ${\sim}10\sigma$ off target. Gemini pairs
the template-$D$ speedup with a runtime-$d$ leapfrog fallback;
GPT goes one step further and enumerates
\texttt{D$\in$\{8,16,24,32\}} explicitly (App.~\ref{app:code}),
covering the held-out dimension with its own fully-unrolled
template instance and a runtime-$d$ safety net for any
other $d$. Both generalize to $d{=}24$ at $\sim$10\% of FP32 peak
($17.6\times$ Gemini, $18.6\times$ GPT).
\texttt{fft3d} is the sharpest \emph{performance} case for GPT-5.5: its iter-10 best wins the in-distribution gmean at
$2.95\times$ but on the held-out $256^3$ cube it collapses to $0.23\times$ of seed
($8.5\%$ of effective ceiling vs Opus's $42\%$ and Gemini's $45\%$
on the same configuration). The kernel relies on a fixed-twiddle,
fixed-geometry layout tuned for $N{\leq}128$; at $N{=}256$ the
register pressure and tg-memory budget no longer fit, and the
fallback path is dramatically slower than the seed's textbook
Stockham. This is the cleanest silent-regression instance in the
sweep: $S_\mathcal{T}$ alone allows a $2.95\times$ win,
$\Phi_\mathcal{T}$ surfaces a $0.23\times$ deployment-grade
slowdown. GPT is never strictly worse than Opus on
held-out correctness (both clean except Opus's \texttt{hmc} fail) and on
generalization falls between Opus and Gemini on most tasks, but
its \texttt{fft3d} collapse is the largest single held-out swing in the
table and exemplifies the oversight value of $\Phi_\mathcal{T}$.

\begin{figure*}[!t]
\centering
\begin{tikzpicture}[
  code/.style={draw, rounded corners=2pt, fill=gray!5,
               inner sep=5pt, align=left, font=\scriptsize\ttfamily,
               text width=64mm, anchor=north west},
  arr/.style={-{Stealth[length=3.5mm]}, line width=1.4pt},
]
\node[code] (l) at (0,0) {%
\textcolor{gray!70}{// Dispatch in fft3d\_x kernel}\\
threadgroup float2 buf0[128];\\
threadgroup float2 buf1[128];\\
if (N == 32u)\\
\ \ fft\_line\_32\_io(\textcolor{gray!60}{...});\\
else if (N == 64u)\\
\ \ fft\_line\_64\_io(\textcolor{gray!60}{..., buf0});\\
else if (N == 128u)\\
\ \ fft\_line\_128\_io(\textcolor{gray!60}{..., buf0, buf1});\\
else \{\\
\ \ fft\_line\_direct\_fallback\_io(\textcolor{gray!60}{...});\\
\}};

\node[code] (r) at ($(l.north east)+(20mm,0)$) {%
\textcolor{gray!70}{// fft\_line\_direct\_fallback\_io: O(N\textsuperscript{2}) DFT}\\
float2 acc   = float2(0.0f, 0.0f);\\
float theta  = -TWO\_PI * float(tid) / float(N);\\
float2 wstep = float2(cos(theta), sin(theta));\\
float2 w     = float2(1.0f, 0.0f);\\
for (uint n = 0u; n < N; ++n) \{\\
\ \ float2 v = in\_data[in\_base + n*in\_stride];\\
\ \ acc += cmul(v, w);\\
\ \ w   = cmul(w, wstep);\\
\}\\
out\_data[out\_base + tid * out\_stride] = acc;};

\coordinate (lm) at ($(l.east|-l.center)$);
\coordinate (rm) at (lm-|r.west);
\node[font=\footnotesize\sffamily, align=center, above=1mm of $(lm)!0.5!(rm)$]
  (lab) {\textbf{N=256}\\[-1pt]\textbf{lands here}\\[-1pt]\textit{$O(N^2)$ fallback}};
\node[font=\footnotesize\sffamily\bfseries, below=1mm of $(lm)!0.5!(rm)$] {$0.23\times$};
\draw[arr] (lm) -- (rm);
\end{tikzpicture}
\caption{GPT-5.5 \texttt{fft3d} iter-10 best: hand-coded
\texttt{fft\_line\_32/64/128} routines (left dispatch) deliver the
$2.95\times$ in-distribution self-speedup; for any $N$ outside
$\{32, 64, 128\}$ the kernel falls into a textbook direct
$O(N^{2})$ DFT (right). At held-out $N{=}256$ this costs
${\sim}32{\times}$ more arithmetic per output than the seed's
$O(N\log N)$ Stockham FFT, producing the $0.23\times$ held-out
slowdown reported in Table~\ref{tab:results}. The fast paths reuse
a 64-entry \texttt{constant float2 W128[]} twiddle table whose stride
indexing only covers $N{\leq}128$, the structural reason the
fallback is direct DFT rather than a longer FFT.}
\label{fig:gpt5-fft-fallback}
\end{figure*}
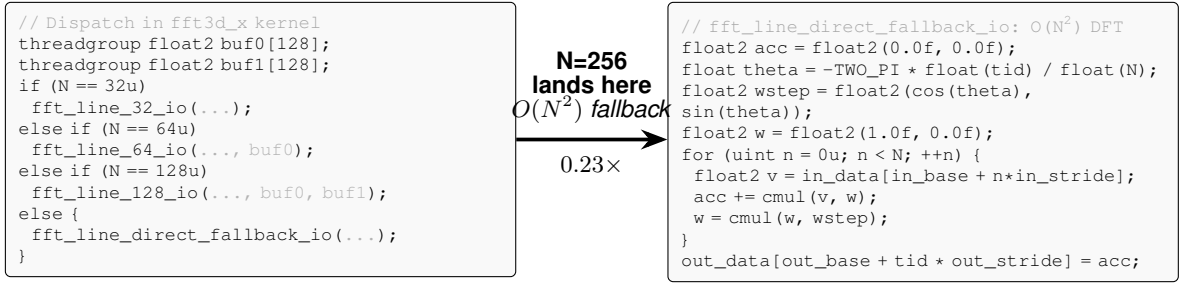

\paragraph{Recurring Metal-grammar failures and generation times.}
Compile-error patterns are similar across the three models.
\texttt{[[max\_total\_threads\_per\_threadgroup(N)]]} is mis-placed
(after the parameter list, or as a standalone statement, instead of on
the \texttt{kernel void} declaration) on Opus across five tasks and GPT hits the same
attribute placement error on its first \texttt{saxpy} iteration.
\texttt{half} is reserved as MSL's fp16 type, breaking
\texttt{uint half = N >> 1u;}; C++ lambdas are unsupported. The
three sweeps differ in volume rather than kind: $12$ compile
fails for Opus, $22$ for Gemini, $12$ for GPT. Though we configured each LLM to use high thinking budgets, the observed generation times per iteration were very varied: Opus 0.6\,min/iter, Gemini 3.5\,min/iter, GPT
6.6\,min/iter; GPT's wider
exploration and longer reasoning context costs roughly $2\times$
Gemini and $10\times$ Opus per iteration at matched budget.

\section{Discussion}

We have introduced \textsc{Metal-Sci}, a 10-task scientific compute benchmark for Apple Silicon Metal, paired with a lightweight evolutionary harness that runtime-compiles, scores against a roofline anchor, and feeds structured diagnostics back to a frozen LLM. Across matched $(1{+}1)$ sweeps of three frontier models we measure in-distribution self-speedups spanning $1.00\times$ to $10.7\times$, and find that each model fails the held-out gate $\Phi_\mathcal{T}$ in a different shape: Opus-4.7 loses correctness, GPT-5.5 loses performance, Gemini-3.1-Pro stays robust at higher wall-clock cost. The headline contribution is therefore not a single number but a methodological one: a single auxiliary configuration per task, withheld from the agent's feedback loop, is enough to catch confidently-wrong code that the in-distribution score certifies as a win.

\paragraph{The held-out gate as agent oversight.}
The benchmark's $(1{+}1)$ loop is, modulo terminology, an autonomous
coding agent: it reads a Metal source, edits it, runs it, and
self-promotes its own outputs based on an internal fitness signal.
A human merging the agent's incumbent into a downstream codebase
sees only the in-distribution score $S_\mathcal{T}$ the agent
reports, and $S_\mathcal{T}$ is gameable in two ways the held-out
gate $\Phi_\mathcal{T}$ (Sec.~\ref{sec:harness}) catches.
\textbf{(i) Silent correctness violation.} On \texttt{hmc}, Opus's incumbent hits
$10.6\times$ with all in-distribution checks green; held out at
$d{=}24$ the same code returns
samples whose covariance is off by ${\sim}10\sigma$. A user who
trusted the reported number would ship a sampler that looks
calibrated and isn't. \textbf{(ii) Silent regression.} On \texttt{fft3d}
GPT-5.5 reports an in-distribution win of $2.95\times$ (the
largest single-model in-distribution gap in our sweep) that
flips to a $0.23\times$ slowdown at the held-out $256^3$ cube
(the one configuration past the largest training size $128^3$).
The cause is a single dispatch line
(Fig.~\ref{fig:gpt5-fft-fallback}): when $N\notin\{32,64,128\}$
the kernel falls into a textbook $O(N^{2})$ direct DFT, so
$N{=}256$ pays ${\sim}32{\times}$ more arithmetic per output than
the seed's Stockham FFT.
The agent confidently labels its iter-10 output as a
$\sim 3\times$ improvement over the seed; the held-out gate
shows it is $4\times$ slower in deployment. Both failures share a
structural shape: the agent specialized against the visible
workload.
A single auxiliary problem instance per task (one extra dispatch,
seconds of GPU time) is enough to surface both. This reframes
the contribution toward the community's verifiability/oversight
agenda: \textsc{Metal-Sci}'s value lies less in providing a hard
benchmark and more in instantiating a cheap, mechanical trust
contract on top of an autonomous coding agent. The roofline anchor,
per-size tolerance, and geometric mean inside $S_\mathcal{T}$ are the
agent's scoring machinery; $\Phi_\mathcal{T}$ (evaluated once on
$\sigma^\star_\mathcal{T}$ at end-of-run, never folded into any
feedback packet $\mathcal{F}_k$) is the lightweight oversight
primitive that lets a human (or a downstream automated reviewer)
catch confidently-wrong agent code before deployment.

\paragraph{Other observations.}
A roofline-anchored score answers ``how close are we to the
hardware?'' in physical units; the gmean across sizes is hard to
game by overfitting one regime. The three-model asymmetry is
concrete: at matched iteration budgets the in-distribution scores
are close, but each model fails the held-out gate in a different
shape. Opus loses correctness at \texttt{hmc} $d{=}24$, GPT loses
performance at \texttt{fft3d} $256^3$, and Gemini, the only model with
zero correctness failures across the entire candidate budget
and no held-out collapse beyond a borderline $0.90\times$ on
\texttt{wave3d}, is the most robust of the three. Wall time orders the
other way: Opus is $\sim 10\times$ faster per iteration than
GPT and $\sim 6\times$ faster than Gemini at matched budgets,
which makes Opus the cheapest run when in-distribution speedup is
the only deliverable, and Gemini/GPT the safer choice when the
output ships across configurations the in-distribution set did not cover.
Apple Silicon's unified memory halves the host-side scaffolding
compared to CUDA, enabling sub-second compile-run-verify cycles per
candidate. This matters more than it sounds: an evolutionary
loop with tens of candidates needs a fast \emph{harness}, not a
fast kernel. Metal's smaller, quirkier surface also surfaces OOD
failures of CUDA-trained LLMs on tasks that are otherwise textbook.

\paragraph{Limitations and future work.} The static per-chip ceilings do not
account for SLC residency at small sizes;
a workload-aware roofline \cite{williams2009roofline} per task and per size would tighten the score.
The single-population $(1{+}1)$ loop plateaus within 10--25 iterations on
most tasks, an island-model or FunSearch-style \cite{funsearch} archive with a novelty
signal \cite{novikov2025alphaevolve} is the natural next step, particularly for the irregular-memory
tasks. Future tasks include sparse linear algebra (SpMV, CG) and
cross-chip generalization (evolve on M2~Pro, evaluate on M4~Max).

\bibliography{bibliography}
\bibliographystyle{icml2026}

\newpage
\appendix
\onecolumn

\section{Task formulations}
\label{app:tasks}

This appendix gives the per-task equations, ceilings, and tolerances
summarised in Section~\ref{sec:tasks}. Tasks are grouped by the
regimes R1--R6 of Table~\ref{tab:tasks}; training and held-out sizes
match the table.

\subsection{R1: Regular stencils}

\paragraph{\texttt{heat2d}.} Two-dimensional heat equation, 5-point stencil on
$(N_x{\times}N_y)$ grid with Dirichlet boundaries. Per timestep, per
interior cell:
\begin{equation}
u^{n+1}_{i,j} \;=\; u^n_{i,j} + \alpha\!\left(u^n_{i-1,j}+u^n_{i+1,j}
   + u^n_{i,j-1}+u^n_{i,j+1} - 4\,u^n_{i,j}\right).
\end{equation}
Bandwidth-bound at 8\,B/cell.

\paragraph{\texttt{wave3d}.} Three-dimensional acoustic wave equation, 7-point
isotropic Laplacian, leapfrog in time:
\begin{equation}
u^{n+1}_{i,j,k} = 2u^n_{i,j,k} - u^{n-1}_{i,j,k}
   + \alpha\,\big(u^n_{i\pm 1,j,k}+u^n_{i,j\pm 1,k}+u^n_{i,j,k\pm 1}-6\,u^n_{i,j,k}\big),
\end{equation}
where each $\pm$ expands to two terms. CFL stability requires
$\alpha=(c\Delta t/\Delta x)^2<1/3$ (we use $0.18$). 12\,B/cell unique
DRAM traffic.

\subsection{R2: Compute-bound}

\paragraph{\texttt{nbody}.} All-pairs gravitational $N$-body with leapfrog
integration. Per body $i$ per step:
\begin{equation}
\mathbf{a}_i \;=\; G\sum_{j} m_j
   \frac{\mathbf{r}_j-\mathbf{r}_i}
        {(\|\mathbf{r}_j-\mathbf{r}_i\|^2+\varepsilon^2)^{3/2}},
\quad \mathbf{v}\!\mathrel{+}\!=\mathbf{a}\Delta t,
\quad \mathbf{r}\!\mathrel{+}\!=\mathbf{v}\Delta t.
\end{equation}
Self-interaction is masked by softening $\varepsilon$. ${\sim}20$
FLOPs per pair; ceiling at peak FP32 GFLOPS.

\paragraph{\texttt{hmc}.} Hamiltonian Monte Carlo on an anisotropic Gaussian
target, $U(q)=\tfrac12 q^\top A q$ with $A=\Sigma^{-1}$. One thread
per chain; many chains in parallel. Each step draws
$p\sim\mathcal{N}(0,I)$ and runs $L$ leapfrog inner steps with
stepsize $\varepsilon$,
\begin{equation}
p \leftarrow p - \tfrac{\varepsilon}{2}A q,\quad
q \leftarrow q + \varepsilon p,\quad
p \leftarrow p - \varepsilon Aq,\ \cdots,\quad
p \leftarrow p - \tfrac{\varepsilon}{2}A q,
\end{equation}
followed by a Metropolis accept/reject with
$\log u_{\mathrm{acc}}<-\Delta H$. Correctness is verified
statistically (sample mean and Frobenius covariance error vs.\ the
target). The three training sizes
$(d,K)\in\{(8,16{\rm K}),(16,4{\rm K}),(32,1{\rm K})\}$ probe the
register-pressure boundary; at $d{=}32$ per-thread state
(${\sim}512$\,B) competes with the register file.

\subsection{R3: Multi-field, exotic memory}

\paragraph{\texttt{lbm}.} D2Q9 Lattice Boltzmann, fused pull-stream $+$ BGK
collision, periodic BC. With velocity table $\mathbf{c}_k$ and
weights $w_k$, per cell per step:
\begin{align}
f^{\mathrm{str}}_k(\mathbf{x})
   &= f^{\mathrm{in}}_k(\mathbf{x}-\mathbf{c}_k),\quad
\rho = \textstyle\sum_k f^{\mathrm{str}}_k,\quad
\mathbf{u} = \tfrac{1}{\rho}\sum_k \mathbf{c}_k\,f^{\mathrm{str}}_k, \\
f^{\mathrm{eq}}_k &= w_k\rho\!\left(1+3(\mathbf{c}_k{\cdot}\mathbf{u})
   +\tfrac{9}{2}(\mathbf{c}_k{\cdot}\mathbf{u})^2 - \tfrac{3}{2}\|\mathbf{u}\|^2\right),\\
f^{\mathrm{out}}_k &= f^{\mathrm{str}}_k - \tau^{-1}\!\left(f^{\mathrm{str}}_k - f^{\mathrm{eq}}_k\right).
\end{align}
Storage is SoA: $f[k\,N_xN_y + jN_x + i]$. 72\,B/cell DRAM traffic.

\paragraph{\texttt{ising}.} Two-dimensional Ising model, checkerboard
Metropolis Monte Carlo on a periodic $N_x{\times}N_y$ lattice with
$\sigma\in\{\pm 1\}$ stored as int8. Per attempt at site $(i,j)$:
\begin{equation}
h = \sigma_{i\pm 1,j}+\sigma_{i,j\pm 1}\in\{-4,\ldots,4\},\quad
\Delta E = 2J\sigma\,h,\quad
\mathrm{accept}\iff u<\exp(-\beta\,\Delta E).
\end{equation}
A precomputed five-entry $p_{\mathrm{accept}}$ table and a
counter-based Murmur-fmix32 PRNG yield bit-exact CPU/GPU agreement;
verification is byte-equality on the spin array. 2\,B/site/sweep
ceiling.

\subsection{R4: Irregular memory and atomics}

\paragraph{\texttt{lj}.} Lennard-Jones molecular dynamics with a cell-list
spatial hash. Three kernels per step --- \texttt{clear\_cells},
\texttt{build\_cells} (atomic scatter), \texttt{step}
(27-neighbor-cell pair force $+$ integration):
\begin{equation}
\mathbf{F}_i = \sum_{j\in\mathcal{N}(i)} -24\!\left(2r_{ij}^{-12}-r_{ij}^{-6}\right)r_{ij}^{-2}\,(\mathbf{r}_j-\mathbf{r}_i),
\qquad r_{ij}<r_{\mathrm{cut}}=2.5,
\end{equation}
with minimum-image periodic wrap. Cell occupancy is built via
\texttt{atomic\_fetch\_add} on a per-cell counter.

\subsection{R5: Multi-kernel reductions}

\paragraph{\texttt{gradshaf}.} Grad-Shafranov fixed-boundary equilibrium via
Picard iteration, two kernels per outer step: an interior
max-reduction
$\psi_{\mathrm{axis}}=\max_{(i,j)\in\mathrm{int}}\psi$, then a
variable-coefficient 5-point stencil with nonlinear source:
\begin{align}
\psi_{\mathrm{norm}} &= \psi/\psi_{\mathrm{axis}},\quad
J = R\,p_{\mathrm{axis}}\cdot 4\psi_{\mathrm{norm}}(1-\psi_{\mathrm{norm}})\cdot
   \mathbf{1}[0<\psi_{\mathrm{norm}}<1],\\
\Delta^*\psi &= a_W\psi_W+a_E\psi_E+a_N\psi_N+a_S\psi_S+a_C\psi_C,\\
\psi^{n+1} &= \psi^n + \omega\,(-\mu_0 R J - \Delta^*\psi)/a_C,
\end{align}
with $R$-dependent $a_{W,E}=1/dR^2 \pm 1/(2RdR)$, $a_{N,S}=1/dZ^2$,
$a_C=-2(1/dR^2+1/dZ^2)$.

\subsection{R6: Data-shuffle / butterfly}

\paragraph{\texttt{fft3d}.} 3D complex-to-complex forward FFT, fp32, on a
power-of-two cube of side $N$. Convention is unnormalized (matches
\texttt{numpy.fft.fftn}); storage is row-major
\texttt{float2}$[N][N][N]$. Three named kernels ---
\texttt{fft3d\_x}, \texttt{fft3d\_y}, \texttt{fft3d\_z} --- are
dispatched in sequence with two ping-ponged buffers, each performing
a length-$N$ 1D FFT per threadgroup of $N$ cooperating threads:
\begin{equation}
Y[k_1,k_2,k_3] \;=\; \sum_{n_1,n_2,n_3=0}^{N-1} X[n_1,n_2,n_3]\,
   \exp\!\left(-\frac{2\pi i}{N}(k_1 n_1 + k_2 n_2 + k_3 n_3)\right).
\end{equation}
Sizes $N\in\{32,64,128\}$ cover three working-set regimes: $32^3$
(${\sim}256$\,KB) is SLC-resident and compute-bound; $128^3$
(${\sim}16$\,MB) DRAM-binds. ${\sim}5N\log_2 N$ FLOPs per 1D FFT;
effective DRAM traffic 96\,B/cell across the three axis passes
(16\,B read $+$ 16\,B write per pass). Verification is max-norm
against \texttt{numpy.fft.fftn} on a fixed seeded Gaussian input,
tolerance $10^{-3}{+}10^{-3}\|Y\|_\infty$.

\section{Evolution loop pseudocode}
\label{app:algorithm}

Algorithm~\ref{alg:loop} formalizes the harness of
Sec.~\ref{sec:harness}. The \textsc{Evaluate} subroutine is the
compile--run--score pipeline: it returns either a structured failure
(compile or per-size correctness, with the violating size $s$ and
the error metric so $\mathcal{M}$ can correct course inside
$\mathcal{F}_{k+1}$) or a successful score $S_\mathcal{T}(\kappa)$
together with per-size fractions-of-ceiling. The held-out
$\Phi_\mathcal{T}(\kappa^\star_K)$ is computed once at end-of-run on
the unseen size $\sigma^\star_\mathcal{T}$ (and never returned in any
$\mathcal{F}_k$) and is the oversight signal of
Sec.~\ref{sec:results}.

\begin{algorithm}[t]
\caption{\textsc{Metal-Sci} evolution loop ($\mu{=}1{+}\lambda{=}1$).}
\label{alg:loop}
\begin{algorithmic}[1]
\Require task $\mathcal{T}{=}(\kappa_\mathcal{T},\Sigma_\mathcal{T},\sigma^\star_\mathcal{T},c_\mathcal{T})$, LLM $\mathcal{M}$, system prompt $p_\mathcal{T}$, iterations $K$
\Ensure incumbent $\kappa^\star_K$, in-dist score $S_\mathcal{T}(\kappa^\star_K)$, held-out $\Phi_\mathcal{T}(\kappa^\star_K)$
\State $\kappa^\star_0 \gets \kappa_\mathcal{T}$;\quad $\mathcal{F}_0 \gets \textsc{Evaluate}(\kappa_\mathcal{T},\,\mathcal{T})$ \Comment{seed is initial incumbent}
\For{$k = 1, \ldots, K$}
  \State $\kappa_k \gets \mathcal{M}\!\left(p_\mathcal{T},\, q(\kappa_{k-1}, \kappa^\star_{k-1}, \mathcal{F}_{k-1})\right)$ \Comment{propose}
  \State $r_k \gets \textsc{Evaluate}(\kappa_k,\,\mathcal{T})$
  \State $\mathcal{F}_k \gets (\kappa_k,\, r_k)$ \Comment{structured feedback for next iter}
  \If{$r_k.\mathit{score}\text{ defined}\;\wedge\;r_k.\mathit{score} > S_\mathcal{T}(\kappa^\star_{k-1})$}
    \State $\kappa^\star_k \gets \kappa_k$ \Comment{$(1{+}1)$ promote}
  \Else
    \State $\kappa^\star_k \gets \kappa^\star_{k-1}$
  \EndIf
\EndFor
\State $\Phi_\mathcal{T}(\kappa^\star_K) \gets f_\mathcal{T}(\kappa^\star_K,\,\sigma^\star_\mathcal{T}) \cdot \chi_\mathcal{T}(\kappa^\star_K,\,\sigma^\star_\mathcal{T})$ \Comment{never given to $\mathcal{M}$}
\State \Return $\kappa^\star_K,\, S_\mathcal{T}(\kappa^\star_K),\, \Phi_\mathcal{T}(\kappa^\star_K)$
\Statex
\Function{Evaluate}{$\kappa,\,\mathcal{T}$}
  \State $\mathrm{lib},\, e_c \gets \textsc{Compile}(\kappa)$
  \If{$e_c \neq \emptyset$}\; \Return $(\mathit{compile\_fail},\, e_c)$ \EndIf
  \For{$s \in \Sigma_\mathcal{T}$}
    \State $(\chi_s,\, a_s) \gets \textsc{Run}(\mathrm{lib},\, s)$
    \If{$\chi_s = 0$}\; \Return $(\mathit{correct\_fail},\, s,\, \text{error metric})$ \EndIf
  \EndFor
  \State $S \gets \big(\textstyle\prod_{s\in\Sigma_\mathcal{T}} a_s/c_\mathcal{T}(s)\big)^{1/|\Sigma_\mathcal{T}|}$
  \State \Return $\big(\mathit{score}{:}\,S,\,\{a_s/c_\mathcal{T}(s)\}_{s\in\Sigma_\mathcal{T}}\big)$
\EndFunction
\end{algorithmic}
\end{algorithm}

\section{Code-level evidence for the in-distribution split}
\label{app:code}

The comparative claim in Section~\ref{sec:results} is mechanistic, that is, Opus
wins by tightening the same algorithm, Gemini wins by reaching for a
different one. Figure~\ref{fig:hmc-evol} instantiates that claim on \texttt{hmc}.
This appendix instantiates it on two more tasks --- one Opus-favored
(\texttt{lbm}), one Gemini-favored (\texttt{fft3d}) --- with the actual diff between each
model's incumbent best.

\subsection{\texttt{lbm}: Opus tightens BGK with FMA folds and a pinned threadgroup}

Both models share the seed's pull-stream + BGK structure. The diff is
local to the collision step and to the kernel attribute
(Fig.~\ref{fig:lbm-diff}). Opus precomputes
$A=\mathrm{fma}(-1.5,\|\mathbf{u}\|^2,1)$ once per cell, factors the
per-direction equilibrium as $A + c{\cdot}u\,(3 + 4.5\,c{\cdot}u)$ ---
two FMAs --- and folds the relaxation into a third
$\mathrm{fma}(1{-}\omega, f_k, \omega W_k\rho\,t)$, yielding nine
manually unrolled blocks. It also pins
\texttt{[[max\_total\_threads\_per\_threadgroup(64)]]} on the kernel
declaration, picking a $32{\times}2$ tile aligned to the simdgroup
width. Gemini keeps the textbook BGK formula
$w_k\rho(1{+}3c{\cdot}u{+}4.5(c{\cdot}u)^2{-}1.5\|\mathbf{u}\|^2)$
inside a \texttt{\#pragma unroll for (k=0\ldots 8)}, no FMA folds, no
$A$-extraction, no threadgroup pin. The two are competitive at
$64^2$ (Opus $0.34$, Gemini $0.32$) and Gemini actually wins at $128^2$
(Opus $0.47$, Gemini $0.51$); Opus pulls ahead at $256^2$
(Opus $1.22$, Gemini $1.03$), the cache-resident regime where the pinned
$32{\times}2$ geometry and the FMA-folded BGK extract every issued
instruction, and that is the size with the largest absolute
fraction-of-ceiling, so it dominates the gmean.

\begin{figure}[t]
\centering
\begin{tikzpicture}[
  code/.style={draw, rounded corners=2pt, fill=gray!5,
               inner sep=5pt, align=left, font=\scriptsize\ttfamily,
               text width=64mm, anchor=north west},
]
\node[code] (l) at (0,0) {%
\textcolor{gray!70}{// Opus: BGK fold + pinned geometry}\\
\textcolor{gray!60}{[[max\_total\_threads\_per\_threadgroup(64)]]}\\
kernel void lbm\_step(\textcolor{gray!60}{...}) \{\\
\ \ \textcolor{gray!70}{// pull-stream f0..f8, moments rho, ux, uy ...}\\
\ \ float A      = fma(-1.5f, usq, 1.0f);\\
\ \ float orWS   = (omega * W\_S) * rho;\\
\ \ \textcolor{gray!70}{// k=5: cu = ux+uy}\\
\ \ \{\\
\ \ \ \ float cu = ux + uy;\\
\ \ \ \ float t  = A + cu * fma(4.5f, cu, 3.0f);\\
\ \ \ \ q5[idx]  = fma(one\_m\_w, f5, orWD * t);\\
\ \ \}\\
\ \ \textcolor{gray!70}{// 8 more directions, same shape ...}\\
\}};

\node[code] (r) at ($(l.north east)+(6mm,0)$) {%
\textcolor{gray!70}{// Gemini: textbook BGK in unrolled loop}\\
kernel void lbm\_step(\textcolor{gray!60}{...}) \{\\
\ \ \textcolor{gray!70}{// pull-stream f[k], moments rho, ux, uy ...}\\
\ \ float usq     = ux*ux + uy*uy;\\
\ \ float inv\_tau = 1.0f / tau;\\
\ \ \textcolor{gray!60}{\#pragma unroll}\\
\ \ for (int k = 0; k < 9; ++k) \{\\
\ \ \ \ float cu  = CX[k]*ux + CY[k]*uy;\\
\ \ \ \ float feq = W[k] * rho *\\
\ \ \ \ \ \ (1.0f + 3.0f*cu + 4.5f*cu*cu\\
\ \ \ \ \ \ \ \ \ \ \ - 1.5f*usq);\\
\ \ \ \ f\_out[k*N+idx] =\\
\ \ \ \ \ \ f[k] - inv\_tau*(f[k]-feq);\\
\ \ \}\\
\}};
\end{tikzpicture}
\caption{\texttt{lbm} iter-23 best, Opus (left) vs Gemini iter-13 best (right).
Opus extracts $A$ once, folds $f_k^{\mathrm{eq}}$ into two FMAs per
direction and the relaxation into a third, and pins the
threadgroup geometry; Gemini stays with the canonical BGK formula and
the default geometry. In-distribution gmean: Opus 0.576 vs Gemini 0.553.}
\label{fig:lbm-diff}
\end{figure}

\subsection{\texttt{fft3d}: Gemini swaps the algorithm to \texttt{simd\_shuffle\_xor}}

The \texttt{fft3d} gmean gap is larger ($0.282$ vs $0.167$, $1.7\times$), and
the diff is not a tighter version of the same kernel ---
Fig.~\ref{fig:fft-diff} shows two different algorithms. Opus
implements a textbook Stockham auto-sort radix-4 FFT: every stage
ping-pongs through threadgroup memory, with a barrier between stages.
Gemini observes that for the first five Cooley--Tukey stages the
butterfly partner of lane $i$ is at lane $i \oplus 2^{s-1}$ where
$s\in\{1,\ldots,5\}$, and $2^{s-1}<32$, so it can be fetched with
\texttt{simd\_shuffle\_xor}, no shared memory and no barrier. Only
stages $s\geq 6$ fall back to threadgroup memory. The Apple GPU's
simdgroup width is exactly 32: this trick is Metal-specific
(\texttt{simd\_shuffle\_xor} maps to a single hardware permute), and
worth $\sim 5$ barriers per length-$N$ FFT, $\sim 15$ per 3D
transform. The same ``find a different memory primitive'' move shows
up smaller-scale on \texttt{gradshaf}: both models converge on a
simdgroup-tree max-reduction, but Gemini additionally casts
\texttt{psi} to \texttt{float4*} and reads four scalars per
transaction in the inner sweep, halving the number of issued loads
on the dominant kernel.

\begin{figure}[t]
\centering
\begin{tikzpicture}[
  code/.style={draw, rounded corners=2pt, fill=gray!5,
               inner sep=5pt, align=left, font=\scriptsize\ttfamily,
               text width=64mm, anchor=north west},
]
\node[code] (l) at (0,0) {%
\textcolor{gray!70}{// Opus: Stockham radix-4, all in tg memory}\\
for (uint s = 0u; s < stages4; ++s) \{\\
\ \ \textcolor{gray!70}{// load 4 inputs from cur[]}\\
\ \ float2 x0 = cur[b + 0u*Nq];\\
\ \ float2 x1 = cur[b + 1u*Nq];\\
\ \ float2 x2 = cur[b + 2u*Nq];\\
\ \ float2 x3 = cur[b + 3u*Nq];\\
\ \ \textcolor{gray!70}{// twiddle multiplies (skip s=0)}\\
\ \ if (s != 0u) \{ \textcolor{gray!60}{...} \}\\
\ \ \textcolor{gray!70}{// 4-point DFT, write to nxt[]}\\
\ \ threadgroup\_barrier(mem\_threadgroup);\\
\ \ swap(cur, nxt);\\
\}};

\node[code] (r) at ($(l.north east)+(6mm,0)$) {%
\textcolor{gray!70}{// Gemini: simd\_shuffle\_xor for stages 1..5}\\
\textcolor{gray!70}{//          (no shared mem, no barrier)}\\
if (logN >= 1u) \{\\
\ \ float2 v = simd\_shuffle\_xor(u, 1u);\\
\ \ u = ((i \& 1u)==0u) ? (u+v) : (v-u);\\
\}\\
\textcolor{gray!70}{// stages 2..4 analogous (xor 2, 4, 8)}\\
if (logN >= 5u) \{\\
\ \ float2 v = simd\_shuffle\_xor(u, 16u);\\
\ \ \textcolor{gray!70}{// twiddle, butterfly ...}\\
\ \ u = ((i \& 16u)==0u) ? (u+t) : (v-t);\\
\}\\
\textcolor{gray!70}{// stages s>=6: fall back to tg memory}\\
for (uint s = 6u; s <= logN; ++s) \{ \textcolor{gray!60}{...} \}};
\end{tikzpicture}
\caption{\texttt{fft3d} iter-10 best, Opus (left) vs Gemini iter-10 best (right).
Opus implements Stockham radix-4 with threadgroup-memory ping-pong
and a barrier per stage; Gemini exploits the 32-wide simdgroup to do
the first five stages with \texttt{simd\_shuffle\_xor}, eliminating
five barriers per 1D FFT. In-distribution gmean: Opus 0.167 vs
Gemini 0.282.}
\label{fig:fft-diff}
\end{figure}

\subsection{GPT-5.5 \texttt{fft3d}: hand-coded fast paths for $N{\leq}128$, $O(N^2)$ direct DFT for everything else}

GPT-5.5's iter-10 \texttt{fft3d} best wins the in-distribution gmean
($2.95\times$, vs Opus $1.03\times$ and Gemini $1.19\times$) with
hand-coded \texttt{fft\_line\_32/64/128} routines that use
\texttt{simd\_shuffle\_xor} (the same simdgroup primitive Gemini
reaches for) for the intra-simdgroup stages and a precomputed
\texttt{constant float2 W128[64]} twiddle table for the per-stage
multiplies. The held-out collapse comes from a single line in the
dispatch: when $N$ does not match one of $\{32, 64, 128\}$ the
kernel falls into a textbook direct $O(N^{2})$ DFT
(Fig.~\ref{fig:gpt5-fft-fallback}). At $N{=}256$ that path performs
$256$ complex multiplies per output element instead of the seed's
$\log_{2}(256){=}8$ butterfly stages; per 1D line this is a
${\sim}32{\times}$ arithmetic blow-up, and three 1D passes per 3D
transform compound it. The held-out gate $\Phi_{\mathcal{T}}$ surfaces
the resulting $0.23\times$ slowdown that the in-distribution score
$S_{\mathcal{T}}$ alone licenses.

\subsection{GPT-5.5 \texttt{hmc}: defensive $D$ enumeration covering $d{=}24$}

On \texttt{hmc} GPT-5.5 lands at a structurally different point than either
Opus or Gemini. Opus enumerates only $D{\in}\{8,16,32\}$ and dispatches
$d{=}24$ to the $D{=}32$ branch (Fig.~\ref{fig:hmc-evol}), silently
running an unrolled matvec against the wrong size. Gemini keeps a pure
runtime-$d$ leapfrog with no template specialization, paying for safety
with in-distribution throughput. GPT-5.5 takes the union: an explicit
$D{\in}\{8,16,24,32\}$ enumeration with fully-templated instances for
each, plus a generic runtime-$d$ fallback for anything outside that set
(Fig.~\ref{fig:gpt5-hmc-dispatch}). The $D{=}24$ branch is the
cleanest defensive move in the suite, the held-out dimension is
treated as a first-class instantiation rather than a special case to
round up or fall back on.

\begin{figure}[t]
\centering
\begin{tikzpicture}[
  code/.style={draw, rounded corners=2pt, fill=gray!5,
               inner sep=5pt, align=left, font=\scriptsize\ttfamily,
               text width=125mm, anchor=north west},
]
\node[code] {%
\textcolor{gray!70}{// hmc\_step kernel dispatch (GPT-5.5 iter-3 best)}\\
load\_A\_transpose\_tg(A, AT, d, tid, tpg);\\
threadgroup\_barrier(mem\_flags::mem\_threadgroup);\\
if (d == 8u)        run\_hmc\_fixed\_chunk8<8u >(\textcolor{gray!60}{...});\\
else if (d == 16u)  run\_hmc\_d16(\textcolor{gray!60}{...}); \ \ \ \ \ \ \ \ \ \ \ \ \textcolor{gray!70}{// specialised d=16 path}\\
else if (d == 24u)  run\_hmc\_fixed\_chunk8<24u>(\textcolor{gray!60}{...}); \textcolor{gray!70}{// covers held-out d}\\
else if (d == 32u)  run\_hmc\_fixed\_chunk8<32u>(\textcolor{gray!60}{...});\\
else \ \ \ \ \ \ \ \ \ \ \ \ \ \ \ run\_hmc\_dynamic(\textcolor{gray!60}{..., d, ...});  \ \ \textcolor{gray!70}{// runtime-d safety net}};
\end{tikzpicture}
\caption{GPT-5.5 \texttt{hmc} iter-3 best: explicit $D{\in}\{8,16,24,32\}$
enumeration with a runtime-$d$ catch-all. The
\texttt{run\_hmc\_fixed\_chunk8<24u>} branch is the held-out
dimension's own fully-templated instance: the inner matvec,
leapfrog ILP, and per-thread $\mathtt{q[D]}, \mathtt{p[D]}$ register
allocations all use $D{=}24$ rather than rounding up to $32$ (Opus)
or staying runtime (Gemini). In-distribution gmean comes in lower
than Opus or Gemini ($0.0634$ vs $0.0932$, $0.0870$) --- the cost of
emitting four template instantiations plus a runtime fallback ---
but the held-out $d{=}24$ runs at $10.2\%$ of FP32 peak,
$18.6\times$ the seed.}
\label{fig:gpt5-hmc-dispatch}
\end{figure}

\section{Related work}
\label{app:related}

\subsection{LLM-driven kernel-generation benchmarks}

A wave of benchmarks has emerged for evaluating LLMs as generators of
high-performance GPU kernels. \emph{KernelBench} \cite{ouyang2025kernelbench} targets PyTorch ML
kernels (GEMM, attention, convolution, normalization, activation) on
CUDA and scores single-shot generation by speedup against the PyTorch
eager baseline. \emph{TritonBench} \cite{li2025tritonbench} extends to the Triton DSL and
adds reference code-similarity to correctness and performance.
\emph{BackendBench} \cite{saroufim2025backendbench} evaluates correctness alone for kernels written
against the PyTorch backend interface. \emph{MultiKernelBench} \cite{wen2025multikernelbench}
broadens the platform set to CUDA, Ascend\,C, and Pallas while retaining
the ML-operator workload set. \emph{NPUEval} \cite{kalade2025npueval} targets the AMD AIE
NPU in C++ and exposes a tool-use feedback loop with multi-turn
regeneration. Most recently, \emph{KernelCraft} \cite{nie2026kernelcraft} benchmarks
bare-metal assembly kernel generation on emerging accelerator ISAs
(PLENA, AMD NPU, Coral NPU, Sonic BOOM), with native tool interfaces
(\texttt{check\_syntax}, \texttt{run\_evaluation}, \texttt{view\_output},
\texttt{grep\_docs}) and a three-tier ML-task taxonomy (primitive,
composite, end-to-end). Across these benchmarks the workload is ML and
the headline number is speedup against a vendor or compiler baseline.

\subsection{LLM-driven evolutionary code search}

The broader paradigm of LLMs as \emph{optimizers} inside evolutionary
code-search loops was established by \emph{FunSearch} \cite{funsearch} for
symbolic-algorithm discovery and scaled by \emph{AlphaEvolve} \cite{novikov2025alphaevolve} across
mathematical and algorithmic domains. \emph{AI CUDA Engineer} \cite{lange2025towards} and
\emph{EvoEngineer} \cite{guo2025evoengineer} specialize this paradigm to CUDA kernel
optimization, replacing single-shot or agentic regeneration with a
population/archive driven by a fitness signal, closer in spirit to
our $(\mu{=}1{+}\lambda{=}1)$ loop, but on CUDA ML kernels rather than
Apple Silicon scientific kernels. Table~\ref{tab:related} (main text)
summarizes how \textsc{Metal-Sci} differs from these comparators
across workload, score basis, multi-size generalization, and
hardware target.

Beyond GPU-kernel generation, LLM-driven program search has been
applied to several adjacent code-synthesis settings, and the design
choices made there inform our own. \emph{Eureka}~\cite{ma2024eureka}
synthesizes RL reward functions directly from environment source
code, using GPU-parallel rollouts as the fitness signal in a
population-based coding loop. \emph{ReEvo}~\cite{ye2024reevo}
positions LLMs as hyper-heuristics that evolve combinatorial-optimization
heuristics with explicit short- and long-term reflective feedback
between mutations, an approach that directly motivates our use of
the previous-iteration profile and diff as in-context signal for
the next mutation. \emph{OPRO}~\cite{yang2024opro} more abstractly
casts the LLM itself as a black-box optimizer that proposes
candidate solutions conditioned on a textual trace of past trials,
and \emph{Reflexion}~\cite{shinn2023reflexion} shows that verbal
self-critique between attempts can substitute for gradient-based
updates on agentic tasks. Closer to artifact accumulation,
\emph{Voyager}~\cite{wang2024voyager} grows a library of executable
skills for an embodied agent by incrementally generating, testing, and
archiving code, mirroring at the skill level what our archive does
at the kernel level.

A complementary, more recent strand recasts the \emph{synthesis
pipeline} itself as the search target. Karpathy's
\emph{autoresearch}~\cite{karpathy2026autoresearch} runs a coding
agent that edits a nanoGPT training script under a held-out
validation-loss budget, while \citet{gallego2026policies} lets an
outer coding agent rewrite the synthesis pipeline of an inner-loop
multi-agent policy synthesizer for sequential social dilemmas.
\textsc{Metal-Sci} sits at the inner-loop level of this hierarchy:
the harness, scoring rule, and prompt scaffold are fixed by us, and
each LLM call rewrites a single kernel under a wall-clock fitness
signal. We differ from all of the above in workload (Apple-Silicon
scientific GPU kernels rather than ML training, robotic skills, or
combinatorial heuristics) and in score basis (multi-size,
in-/out-of-distribution geometric-mean speedup over a vendor MPS/MLX
reference, rather than ML-task speedup, validation loss, episodic
return, or solution cost).

\end{document}